\documentclass{article}

     \usepackage[preprint]{neurips_2019}

\usepackage{amsmath}
\usepackage{microtype}
\usepackage{graphicx}
\usepackage{subfigure}

\usepackage{hyperref}

\usepackage[utf8]{inputenc} 
\usepackage[T1]{fontenc}    
\usepackage{hyperref}       
\usepackage{url}            
\usepackage{booktabs}       
\usepackage{amsfonts}       
\usepackage{nicefrac}       
\usepackage{microtype}      

\title{The role of a layer in deep neural networks: a Gaussian Process perspective}

\author{%
  Oded Ben-David and Zohar Ringel \\
  Racah Institute of Physics, \\
  Hebrew University of Jerusalem, \\
  Jerusalem 91904, Israel.  \\
  \texttt{zohar.ringel@mail.huji.ac.il} \\
}

\begin{document}

\maketitle

\begin{abstract}
  A fundamental question in deep learning concerns the role played by individual layers in a deep neural network (DNN) and the transferable properties of the data representations which they learn. To the extent that layers have clear roles one should be able to optimize them separately using layer-wise loss functions. Such loss functions would describe what is the set of good data representations at each depth of the network and provide a target for layer-wise greedy optimization (LEGO). Here we derive a novel correspondence between Gaussian Processes and SGD trained deep neural networks. Leveraging this correspondence, we derive the Deep Gaussian Layer-wise loss functions (DGLs) which, we believe, are the first supervised layer-wise loss functions which are both explicit and competitive in terms of accuracy. Being highly structured and symmetric, the DGLs provide a promising analytic route to understanding the internal representations generated by DNNs. 
\end{abstract}

\section{Introduction}

\label{Introduction}
Several pleasant features underlay the success of deep learning: The scarcity of bad minima encountered in their optimization [\cite{Draxler2018,Choromanska2014}], their ability to generalize well despite being heavily over-parametrized [\cite{Neyshabur2018,Neyshabur2014}] and expressive [\cite{Zhang2016}], and their ability to generate internal representations which generalize across different domains and tasks [\cite{Yosinski:2014,Sermanet2013}]. Our current understanding of these features is however largely empirical. 


Internal representation are a key ingredient in transfer learning [\cite{Yosinski:2014,Sermanet2013}] (and some semi-supervised learning schemes [\cite{Kingma2014})]. Here one trains a DNN on task $A$ with large amounts data (say image classification) cutting and freezing several of the lowest layers of that DNN, adding a smaller DNN on top the these frozen layers, and training it for task $B$ (say localization of objects in images) with a smaller dataset. The fact that transfer learning often works quite well implies that, to some degree, layers in a DNN learn data representations which are "useful", or aid in solving the task, even without knowledge on the particular weights of subsequent layers. 

A way of formalizing usefulness of internal representations is to consider layer-wise greedy optimization. Indeed, if weights-specific knowledge of subsequent layers is unimportant, it should be possible to train each layer of the DNN individually. Such optimization should use, at most, knowledge about the architecture of subsequent layers and the task at hand. Thus good internal-representations should miminize a layer-wise loss functions that depends, at most, on the architecture and the task. Such loss functions would also allow one to determine whether a layer can be successfully transferred by measuring how well these representations score on the layer-wise loss functiosn associated with different architectures and tasks. The ability to draw analytic insights from such layer-wise loss functions depends heavily on how accurate and explicit they are. 

{\bf Our contributions:} 1. We derive a correspondence between full batch training of wide DNNs with added white-noise and Gaussian Processes. We argue that this correspondence, along with a previous one that applies at zero noise [\cite{Jacot2018}], give two different limits where our loss-functions becomes optimal. 2. We leverage this correspondence to derive a novel set of explicit supervised layer-wise loss functions, Deep Gaussian Layer-wise losses (DGLs), and test them on fully connected DNNs. The DGLs lead to state-of-the-art performance on MNIST and CIFAR10 when used in LEGO and can also be used to monitor standard end-to-end optimization. The DGLs are architecture dependent but only through a few effective parameters.


{\bf Related work:} The idea of analyzing DNNs layer by layer has a long history. Several early successes of deep networks were obtained using LEGO strategies. In particular good generative models of hand-written digits [\cite{Hinton2006}] and phonetics classifiers [\cite{Mohamed2012}] were trained using an unsupervised (i.e. label unaware) LEGO strategy which for the latter work was supplemented by stochastic gradient descent (SGD) fine-tuning. Following some attempts to perform supervised LEGO [\cite{Bengio2006}], the common practice became to use LEGO as a pre-training initialization protocol \cite{LeCun2015}. As simpler initialization protocols came alone [\cite{glorot2010}] SGD on the entire network (end-to-end) became the common practice. More recent works include several implicit loss function based on IB all having in common that an auxiliary DNN has to be trained in order to evaluate the loss [\cite{Bengio2014,elad2019the}]. 

Considering explicit layer-wise greedy training. \citet{Kadmon2016,Meir1988} considered an unsupervised LEGO training algorithm followed by a classifier for datasets resembling Gaussian mixtures. \citet{Kulkarni2017} have used a batch-norm followed by geometric kernel similarity criterion between activations with similar labels to train depth one and two networks and got results 1-3\% below SGD SOTA on CIFAR10 and 2-3\% below SGD SOTA on MNIST. \citet{Krotov2019} used a Hebbian algorithm and got 50\% accuracy on full CIFAR10 compared to our 53.8\%.  While interesting, both methods appear as an alternative to standard training rather than an attempt to understand standard end-to-end training. They also do not generalize in any obvious manner to different architectures, in particular CNNs. In fact we argue below that the layer-wise loss of \citet{Kulkarni2017} as well as \citet{Wilson2015}, are closely related to our DGL loss for specific architectures.  

Turning to correspondences between DNNs and Gaussian Processes. The mapping between wide neural networks with random independently drawn weights and Gaussian Processes (NNGPs) [\cite{Neal1996,Saul2009,Matthews2018}] was recently used to perform Bayesian inference (tractable for Gaussian Processes [\cite{Rasmussen2005}]) on various fully-connected and CNN architectures [\cite{Lee2018,Novak2018,Lee2019}] yielding SOTA on CIFAR-10 and MNIST. Furthermore the prediction of full-batch gradient descent on wide networks, when averaged over initializations, was shown to follow Bayesian Inference on a modified NNGP known as the Neural Tangent Kernel (NTK) [\cite{Jacot2018}]. A closely related earlier work [\cite{Daniely2016}] showed that the empirical correlations between outputs (kernel) of a wide network follow a Gaussian Processes and remain almost constant during training. Notably one may worry that infinitely wide DNNs will be of little use due to over-fitting, however that is not the case in practice. In fact various works show that the wider the network, the better it seems to generalize [\cite{Neyshabur2018,Neyshabur2014}]. Interestingly there is also evidence that GP predictions remain a good approximation even for networks of depth 10000 [\cite{Xiao2018}] at initialization.

\section{Background on Gaussian Processes and Feynman-path-integrals} 
\label{Sec:GPs}
Gaussian Processes are a generalization of multi-variable Gaussian distributions to a distribution of functions ($f(x)$) [\cite{Rasmussen2005}]. Being Gaussian they are completely defined by their first and second moments. The first is typically taken to be zero and second is known as the covariance function ($K_{xx'} = {\rm E}[f(x) f(x')]$, where ${\rm E}$ denote expectation under the GP distribution). 

There are two related but different correspondences between GPs and DNNs which we now describe. First consider a DNN with input $x$ and output $z_W(x)$, where $W$ denotes the set of all weights and biases. Treating $W$ as a random variable with some prior ($P_0(W)$), $z_W(x)$ endows a probability distribution on function space via $f(x)=z_W(x)$. Interestingly in the infinite width (channel) limit, fully-connected (convolutional) DNNs with uncorrelated $P_0(W)$ generate a GP distribution ($P_0[f]$) for $f$ [\cite{Saul2009,Novak2018}]. Such GPs are known as NNGPs. 

To make the relation between DNNs and NNGPs more explicit, it is convenient to use formalism common in physics known as Feynman-path-integrals [\cite{schulman1996}]. Path-integrals allows integrations over the space of function ($\int Df$) as well as delta-functions on the space of functions ($\int Df \delta(f-f_0) f(x_0)...f(x_n) = f_0(x_0)...f_0(x_n)$). The following limit procedure underlays this formalism: one first discretize $x$ very finely. As a result all the integrals and delta-functions over function space become a long yet finite product of standard multi-dimensional integrals and delta-functions. At the end of the computation one takes this discreterization to be infinitely fine. 

Using the path-integral formalism, NNGPs and their covariance-functions can be defined as 
\begin{align}
P_0[f] &= \int dW P_0[W] \delta(f-z_W) \\ \nonumber 
K_{xx'} &= \int Df P_0[f] f(x) f(x') = \int Df dW P_0[W]\delta(f-z_W) f(x)f(x') = \int dW P_0[W] z_W(x)z_W(x')
\end{align}
The first equation here gives a direct interpretation of the NNGP distribution ($P_0[f]$) as describing the probability distribution induced by the network at initialization.  Notably $K_{xx'}$ can be calculated analytically for many activation functions \cite{Saul2009}. Furthermore Bayesian Inference on NNGP is possible \cite{Lee2018,Saul2009} and explicitly given by 
\begin{align}
\label{Eq:GPPred}
l_* &= \sum_{nm} K(x_*,x_n) [ K({\rm D}) + \sigma^2 I]^{-1}_{nm} l_m 
\end{align}
where $x_*$ is a new datapoint, $l_*$ is the target vector typically chosen as a one-hot encoding of the categorical label, $l_m$ are the training targets, $x_n$ are the training data-points, $[K({\rm D})]_{nm} = K_{x_n,x_m}$ is the covariance-matrix (the covariance-function projected on the training dataset (${\rm D}$)), $\sigma^2$ is a regulator corresponding to a noisy measurement of $z_W(x)$, and $I$ is the identity matrix. Some intuition for this formula can be gained by verifying that $x_* = x_q$ yields $l_* = l_q$ when $\sigma^2=0$. Implicit in the above matrix-inversion is the full Bayesian integration ($\int dW$) over all DNNs weights, weighted by their likelihood given the dataset ($P(W|{\rm D})$). 

 
The second correspondence with GPs concerns full-batch gradient descent at vanishing learning rate. It was recently shown analytically that the prediction of a wide networks (or convolutional networks with many channels), follow Eq. \ref{Eq:GPPred} however with a different covariance-function ($\Theta_{xx'}$) related to $K_{xx'}$, known as the Neural Tangent Kernel (NTK).  

\section{From SGD with noise to Bayesian Inference on GPs} 
Here we describe a limit in which SGD can be mapped to Bayesian inference on an NNGP. Consider GD (full-batch SGD) with added white noise in the limit of vanishing learning rate. For sufficiently small learning rate and making the reasonable assumption that the gradients of the loss are globally Lifshitz, the SGD equations are ergodic and converge to the same invariant measure (equilibrium distribution) as the following Langevin equation [\cite{MATTINGLY2002,risken1996fokker}]
\begin{align}
\gamma \frac{d w_i}{dt} &= -m\frac{d^2 w_i}{(dt)^2}-\partial_{w_i}\left(L[z_W]+\sum_i w_i^2/\sigma^2_w\right) + \sqrt{2 \gamma T}\xi_i(t) 
\end{align}
where $m$ (the mass) controls the momentum, $\gamma$ being inversely proportional to the learning-rate, $\xi_i(t)$ being a set of Gaussian white noise ($\langle \xi_i(t) \xi_j(t')\rangle = \delta_{ij} \delta(t-t')$), $T$ accounting for the strength of the noise, and $w_i$ being the set of networks parameters ($W$). The equilibrium-distribution or invariant-measure describing the steady state of the above equation is the Boltzmann distribution [\cite{risken1996fokker}] $P[W] \propto e^{-\frac{L[Z_W]+\sum_i w_i^2/\sigma^2_w}{T}}$. Notably various works argue that at low learning rates, SGD reaches the above equilibrium, approximately [\cite{Mandt2017,Welling2011,Teh2016}], thus extending the practical implications of our derivation. 

We next turn to describe the Langevin dynamics in function space ($f$). The motivation is that $P[W]$ is likely to be highly multi-modal, as many different weights lead to the same $f(x)=z_W(x)$ and therefore the same loss. Thus $P[f]=\int dW P[W] \delta(f-z_W)$ largely mods-out this multi-modality. Using the path-integral formalism and the Boltzmann distribution, the time-averaged prediction of the trained network $l_*$ for an unseen data-point $x_*$ is re-expressed as    
\begin{align}
\label{Eq:LangPred}
l_* &= \int dW P(W) z_W(x_*) = \int Df \int dW z_W(x_*) e^{-\sum_i \frac{w_i^2}{\sigma^2_w T}} \delta(f-z_W) e^{-\frac{L[z_W]}{T}} \\ \nonumber 
&= \int Df f(x_*)e^{-\frac{L[f]}{T}} \int dW e^{-\sum_i \frac{w_i^2}{\sigma^2_w T}} \delta(f-z_W) = \int Df f(x_*) e^{-\frac{L[f]}{T}} P_0[f] \\ \nonumber 
P_0[f] &\propto \int dW e^{-\sum_i \frac{w_i^2}{\sigma^2_w T}} \delta(f-z_W)
\end{align}
where we identify $P_0[W] \propto e^{-\sum_i \frac{w_i^2}{\sigma^2_w T}}$. 


Next taking $L[f]$ to have an MSE form $L[f] = \sum_n (f(x_n)-y_n)^2$ (where $x_n$ and $y_n$ are the data-points and targets) one can now use the standard correspondence between RKHS-regression with an L2 norm and Bayesian inference on GPs (see [\cite{Rasmussen2005}] Ch. 6) to show that Eq. \ref{Eq:LangPred} and \ref{Eq:GPPred} are equal with $\sigma^2 = T$. 

While the above shows that the infinite-time-average of the Langevin dynamics is given by Eq. \ref{Eq:GPPred}, it is unclear how it relates to time averaging over a standard SGD run-time, where one waits until the validation loss equilibrates (loss-relaxation-time). Indeed to make predictions on specific weights, various ergodicity/local-minima issues [\cite{Draxler2018,Choromanska2014}] are likely to imply that far longer time averages, related to the travel time between local minima. 

Notwithstanding we argue that in the wide-network/many-channels regime (in particular with more weights than data-points), and as far as $f(x)$-related quantities are concerned: averaging over several epochs after the loss-relaxation-time should be enough to accurately sample from \ref{Eq:LangPred}. Indeed in such regimes [\cite{Neyshabur2014,Neyshabur2018}] one finds that the different local-minima in $W$ yield similar performance [\cite{Draxler2018,Choromanska2014}] and hence similar $f$'s (see also [\cite{Lee2019}], Fig. S1). Assuming this equivalence of local minima in terms of $f$ then $l*$, being is a function of $f(x)$ alone, wouldn't suffer from ergodicity/local-minima issues. 

As the network becomes narrower and enters the under-parameterized regime, an equivalence of local-minima is unlikely. Still one can consider $M$ different parallel GD+noise processes having different initial conditions. Each of these $M$ processes is likely to settle to a different local-minima, and $l*$ averaged over the loss-relaxation-time and over all $M$ processes should approach the infinite-time average as $M\rightarrow 0$. Thus Eq. (\ref{Eq:GPPred}) can be understood as describing the prediction of the network averaged over both the loss relaxation time and $M$ initial conditions.

\section{Deriving the Deep Gaussian Layer-wise Loss functions} 
\begin{figure}[h]
	\centering
	\includegraphics*[width=16cm,trim=50 580 130 60,clip]{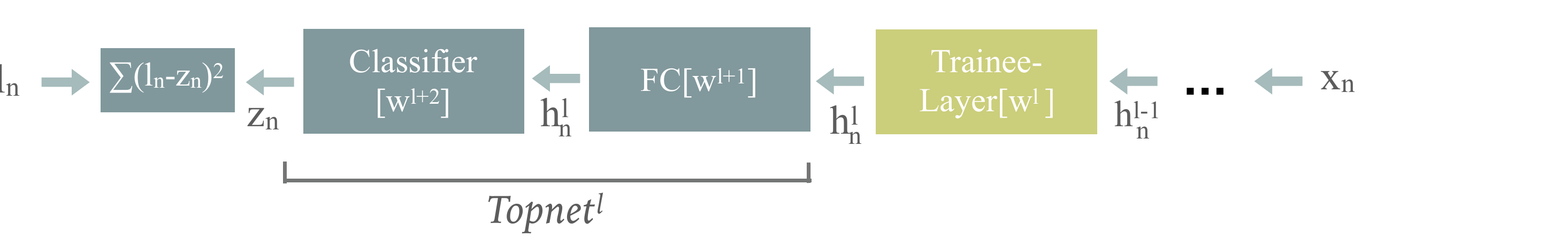}
	\caption{Layer-wise greedy optimization (LEGO). Given an input $x_n$ and larget $l_n$, the $l$'th trainee layer, with weights $w^l$, acts on the output of the $l-1$ layer ($h^{l-1}_n$) and output $h^l_n$, which is $w^l$ dependent. In LEGO we seek to optimize $w^l$ such that $\{(h^l_n,l_n)\}^N_n=1$ is well classified by the $l$'th top-network ($Topnet^l$). Our strategy here is to evaluate this performance using a correspondence between SGD and Bayesian inference on the Gaussian Process associated with this top-network. \label{Arch}} 
\end{figure}

To derive the DGL functions let us start with a LEGO strategy which should be optimal in terms of performance yet highly non-explicit (See Fig.\ref{Arch}): We begin from the input layer and consider it as our current trainee layer ($L_{w^0}$). For every set of its parameters ($w^0$) we perform standard end-to-end training of the entire network between the trainee layer and the classifier (the top-network) with $w^0$ kept frozen. Next we repeat this training infinitely many times and treat the average performance ($Per(w^0)$) as a loss function for the trainee layer. Then we optimize the parameters such that $w^0_{opt} = argmin Per(w^0)$. Subsequently we act on the dataset using $L_{w^0_{opt}}$ to obtain the representation of the dataset in activation space ($h^0_n$). We then repeat the process for the $l=1$ layer with $h^0_n$ as inputs. This process continues until $l=L-1$. The last classifier layer is then trained using MSE Loss. 

Provided that freezing the parameters of the trainee-layer does not induce optimization issues in the top-network SGD, the above procedure is guaranteed to yield the same performance as average end-to-end SGD. Such optimization issues in the top-network, more well known as co-adaptation issues [\cite{Yosinski:2014}], arise from tight coupling between top-network and trainee layer weights. They imply that the trainee layer representations, learned by standard end-to-end training, is highly correlated with the top-network and thus inadequate for transfer learning. 

Co-adaptation is considered adversarial to learning also outside the scope of LEGO and transfer learning. Indeed the success of dropout regularization is partially associated with its ability to mitigate co-adaptation [\cite{Srivastava2014}]. Additionally, co-adaptation being a bad local minima issue, is more likely to occur away from the over-parametrized regime where modern practical interest lies. We thus make an assumption which is that co-adaptation effects are small. Note that if co-adaptation is unavoidable one may still group the co-adapting layers into a block of layers and treat this block as an effective layer in the algorithm discussed below. 

At this point we can derive the DGL loss function based on the following two assumptions: 1. That the exact correspondence between Bayesian inference on NNGP and GD+noise, is a good approximation to standard SGD. 2. That co-adaptation effects are negligible. Given that, $Per[w^l]$ can evaluated using Eq. \ref{Eq:GPPred} and used as our loss. Notably if we plan to train the top-networks using SGD with small learning rates noise it is more appropriate to use the NTK covariance function ($\Theta_{xx'}$) whereas for larger learning rates and with added noise, it is more appropriate to use the NNGP covariance function ($K_{xx'}$). In practice we shall see that both work very well. To this end let us either consider a regression problem with data-point ($h^l_n$) and targets ($l_n$) or rephrase a classification problem as a regression problem by taking $l_n$ to be a one-hot encoding of the categorical labels. For concreteness we focus on the bottom/input layer (see Fig. \ref{Arch}) which acts on the $x_n$ and maps each to its value in the activation space of the input layer ($h^0_n$). Consider training the top-network on the dataset represented by ($h^0_n$, $l_n$). Taking the GP approximation we consider Eq. (\ref{Eq:GPPred}) with $x$ replaced by $h$ and $K_{xx'}$ replaced by $K_{hh'}$ (the covariance-function of the top-network). The resulting equation now describes how an unseen activation $h_*$ would be classified ($l_*$) by a trained top-network. To make this into a loss function for the training dataset, rather than for an unseen point, we adopt a leave-one-out cross-validation strategy: We iterate over all data-points, take each one out in turn, treat it is an unseen point, and measure how well we predict its label using the mean Bayesian NNGP prediction.




Assuming $K_{nm}$ has no zero eigenvalues, taking $\sigma \rightarrow 0$, and performing some straightforward algebra (see App. I.) the MSE loss of the leave-one-out predictions can be expressed using the inverse of $[K(D)]_{nm}$ over the training dataset ($B_{nm} = [K]^{-1}_{nm}$) 
\begin{align}
\label{DGLloss} 
L_{DGL} &= -\sum_{n,m,q} (l_m \cdot l_n) S_{nm}  \\ \nonumber 
S_{nm} &= -\sum_{q} \frac{B_{nq} B_{qm}}{B^2_{qq}}
\end{align}
A few technical comments are in order. The DGL is a function of the trainee layer's parameters via $h^l_n(w^{0})$ which enter $K_{nm}$ whose inverse is $B_{nm}$. Apart from the need to determine the top-networks effective parameters ($\sigma^{2(l)}_{w/b}$) numerically or through meta-optimization, the DGL is an explicit function of all the points in the dataset. We stress that this loss gives a score to a full dataset rather than to points in the dataset. 

We turn to discuss the structure and symmetries of $L_{DGL}$. As $L_{DGL}$ depends on $h^l_n$ only through $B_{qp}$ which in turn depends only on $K(h^l_q,h^l_p)$, it inherits all the symmetries of the latter. For fully connected top-networks it is thus invariant under any orthogonal transformation ($O(d^l)$, where $d^l$ is the dimension of the vector ($h^l_n$) of the $l$'th layer-representation ($h^l_n$). An additional structure is that $L_{DGL}$ depends on the targets only through the dot-product of the targets which, for the one-hot encoded case, means it is zero unless the labels are equal. The $B_{pq}$-dependent central piece ($S_{nm}$) is ``unsupervised" or unaware of the labels. It is a negative definite matrix ensuring that the optimal DGL is zero as one expects from a proxy to the MSE loss. One can think of $S_{nm}$ the sample-similarity measure of the DNN (more specifically the top-network): when $S_{nm}$ is small (large) for two data-points the networks tends to associate different (similar) targets to them in any classification task. Crucially $S_{nm}$ is not a simple pairwise dependence on $h^l_n,h^l_m$, but rather depends on the entire dataset through the covariance-matrix inversion. The DGL function can thus be interpreted as the sample similarity (in the context of the dataset) weighted by the fixed-target similarity.

\section{DGL for the preclassifier layer}
It is illustrative to demonstrate our approach on a case where the inversion of the covariance-matrix can be carried out explicitly. To this end we consider a DNN consisting of a fully-connected or convolutional bottom/input layer ($L_{w^0}$) with weights $w^0$ and any type of activation. This layer outputs a $d$ dimensional activation vector ($h$) which is fed into a linear layer with two outputs $\vec{z} = z_1(x),z_2(x)$. We consider a binary regression task with two targets $\vec{z}= \{(1,0),(0,1)\}$ which can also be thought of as a binary classification task.  


To express the DGL function for this input layer ($L_{DGL@Linear}$), our first task is to find the covariance-function of the top-network namely, the linear layer. Assuming a Gaussian prior of variance $\sigma^2_w$ on each of the linear layer's matrix weights and zero bias it is easy to show (App. II.) that 
\begin{align}
K_{hh'} &= \sigma^2_w h \cdot h' \\ \nonumber 
K_{nm} &= \sigma^2_w H H^T 
\end{align}
where $H$ is an $N$ by $d$ matrix given by $[H]_{ni} = [h^0_n]_i$. 

To facilitate the analysis we next make the reasonable assumption that the number of data-points ($N$) is much larger than the number of labels and also take a vanishing regulator ($\sigma \rightarrow 0$). As a result we find that the covariance-matrix has a kernel whose dimension is at least $N-d \gg d$. To leading order in $d/N$ one finds that $K^{-1} = (\sigma^2_w \sigma)^{-1}P_{K}$, where $P_K$ is the projector onto the kernel of $K_{nm}$. This projector is given by (see App. 3)
\begin{align}
P_K &= (I - H \Sigma^{-1}H^T) \\ \nonumber 
\Sigma &= H^T H 
\end{align}
Indeed one can easily verify that $P_K^2 = P_K$ and that $P_K H H^T=HH^T P_K=0$ as required. Plugging these results into Eq. \ref{DGLloss} one finds that to leading order in $N/d$
\begin{align}
\label{DGLloss_Linear} 
L_{DGL@Linear} &= \sum_{n}|l_n|^2 -\sum_{nm} (l_n \cdot l_m) [H \Sigma^{-1}H^T]_{nm} 
\end{align}

The above equation tell us how to train a layer whose output ($H$) gets fed into a linear classifier. Let us first discuss its symmetry properties. The first term in this equation is constant under the optimization of $w^0$ hence we may discard it. The second term is invariant under any rotation ($O$) of the dataset in activation space ($H \rightarrow H A$). Indeed such transformations can be carried by the classifier itself and hence such changes to the dataset should not affect the performance of the classifier. A bit unexpected is that $L_{DGL@Linear}$ is also invariant under the bigger group of invertible linear transformation ($GL(d)$). While a generic classifier can indeed undo any linear transformation, the prior we put on its weights limits the extent to which it can undo a transformation with vanishing eigenvalues. This enhanced symmetry is a result of taking the $\sigma \rightarrow 0$ limit, which allows the Gaussian Process to distinguish vanishingly small difference in $z(x_n)$. In practice finite $\sigma$ is often needed for numerical stability and this breaks the $GL(d)$ symmetry down to an $O(d)$ symmetry.  

Next we discuss how $L_{DGL@Linear}$ sees the geometry of the dataset. Notably $\Sigma$ is the covariance matrix of the dataset in activation space. Since it is positive definite we can write $\Sigma = \sqrt{\Sigma} \sqrt{\Sigma}$, $\Sigma^{-1} = \sqrt{\Sigma^{-1}} \sqrt{\Sigma^{-1}}$ and therefore $H \Sigma^{-1}H^T = H \sqrt{\Sigma^{-1}} \sqrt{\Sigma^{-1}}H^T$. We then define $\tilde{H} = H \sqrt{\Sigma^{-1}}$ as the normalized dataset. Indeed its covariance matrix ($\tilde{H}^T \tilde{H}$) is the identity.Thus, $L_{DGL@Linear} = const - \sum_{nm} l_n \cdot l_m \tilde{H}_n \tilde{H}_m$. In these coordinates the loss is a simple pairwise interaction between {\it normalized datapoints} which tends to make points with equal (opposite) labels closer (far-apart). This combination of normalization followed encouraging $h_n,h_{m}$, with similar labels to have $h_n \cdot h_m$, is very similar to the loss used in \citet{Kulkarni2017}, such normalization is explicitly performed followed by measure the similarity of $e^{(1-\tilde{h_n}\cdot \tilde{h_m})/(2\sigma^{2})}l_n \cdot l_m$. In that light are results can be viewed as a generalization of those of \citet{Kulkarni2017}, where the somewhat adhoc, $e^{(1-\tilde{h_n}\cdot \tilde{h_m})/(2\sigma^{2})}$ similarity-measure, is replaced by a similarity measure induced by the DNN itself. 



\section{Numerical experiments}
Here we report several numerical experiments aimed at corroborating our analytical predictions. Experiments were conducted on three datasets: MNIST with 10k training samples randomly selected from the full MNIST training set and balanced to have an equal number of samples from each label ($\text{MNIST}_{10k}$), CIFAR10 with 10k training samples similarly selected and balanced in terms of labels ($\text{CIFAR10}_{10k}$). Binary MNIST with only the digits 1 and 7 and 2k training samples ($\text{BMNIST}_{2k}$), similarly selected and balanced in terms of labels. For each dataset, an additional validation set of size equal to the training set, was randomly selected from the full respective training set, excluding the samples selected for the training set. The validation set was balanced in terms of labels. For $\text{MNIST}_{10k}$ and $\text{CIFAR10}_{10k}$ the reported test set was the respective standard test-set and for $\text{BMNIST}_{2k}$ the reported test set was the samples from the standard test-set with labels 1 and 7. The test sets were not balanced in terms of labels. 

All experiments were conducted using fully-connected DNNs, with depth $L$, consisting of $L$ activated layers with fixed width ($d$) and a linear classifier layer with output dimension given by the number of classes. The targets were zero-mean one-hot encoded in all experiments except for $\text{CIFAR10}_{10k}$, where the labels were one-hot encoded. The loss function for all non-DGL training was MSE loss.

For each dataset we conducted the following procedure:
1. End-to-end SGD training under MSE loss
2. Evaluation of the mean-field covariance function of the end-to-end-trained network
3. DGL-Monitored end-to-end SGD training under MSE loss with the same hyperparameters as in step 1. and with the mean-field covariance function evaluated at step 2.
4. LEGO training of all activated layers under DGL, using the mean-field covariance function evaluated at step 2. The activated layers were optimized sequentially, starting from the inputs layer. Each layer was optimized once, then kept frozen during the optimization of subsequent layers.
5. Training of the linear classifier layer only, under MSE loss, with the activated layers frozen, either at the DGL-optimized weights or at the randomly-initialized values.

End-to-end training was done using either vanilla SGD optimizer ($\text{BMNIST}_{2k}$) or Adam optimizer ($\text{CIFAR10}_{10k}$, $\text{MNIST}_{10k}$) with standard internal parameter. All DGL training was done using the Adam optimizer with standard internal parameters. All training was done with fixed learning rates $lr$ and weight decay, $wd$. $lr$ and $wd$ were manually selected for each step in each dataset. The best hyper parameters for each step were selected for minimal loss on the validation set.

\begin{figure}[h]
	\centering
	\includegraphics*[width=7.1cm,trim=0 55 0 0,clip]{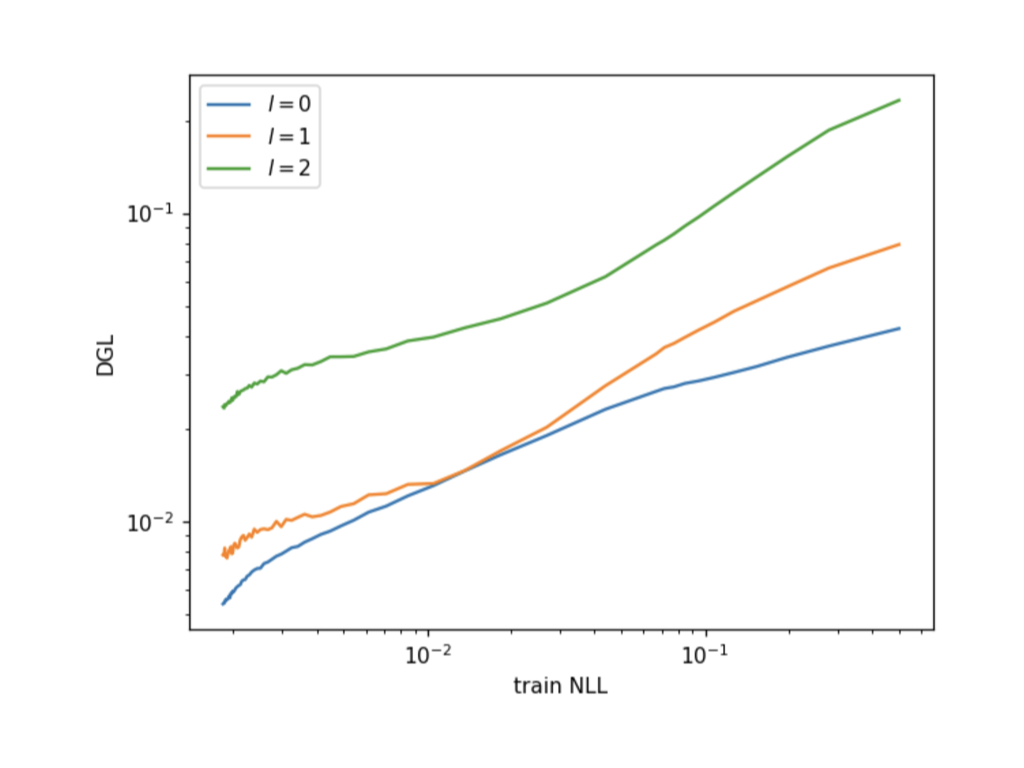}
	\includegraphics*[width=6.7cm,trim=0 195 0 0,clip]{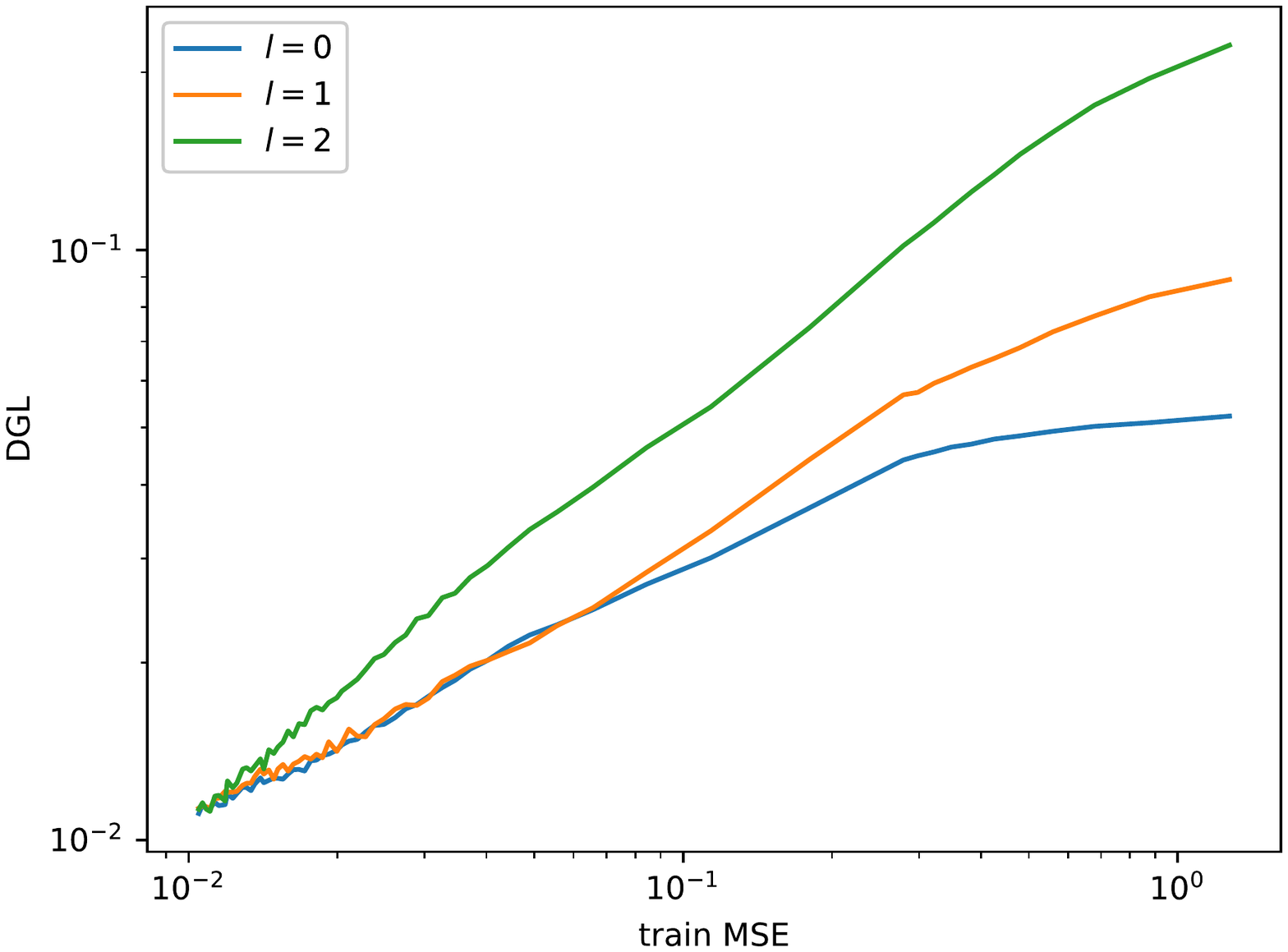}
	\caption{Monitoring end-to-end SGD training using DGL for $\text{BMINST}_{2k}$ with 3 activated layers of width 20. A monotonic behavior of DGLs versus training MSE loss is apparent. Notably all layers converged to similar DGL values which are close to the training values. In addition the ordering of DGLs for the layers is ascending as expected. See code in  https://github.com/dglfunction/DGL\label{Monitoring}} 
\end{figure}

{\bf DGL Monitoring.} Figure (\ref{Monitoring}) shows DGL monitoring (e.g. measuring the DGLs) during standard end-to-end training (step 2.) of a network with $(L,d)=(3,20)$. Even at this small width, DGL tracks end-to-end training very well both for MSE loss and NLL loss, although for NLL the pre-classifier ($l=2$) layer seems somewhat off. It is natural that higher layers would be more affected by the choice of loss. 

{\bf DGL LEGO.} Table \ref{TestResults} shows the test performances of steps 4.\& 5., on the three aforementioned datasets for several $L$ and $d$ choices. End-to-end test accuracy is taken from Ref. [\cite{Lee2018}] apart from $\text{BMNIST}_{2k}$ and the first $\text{CIFAR10}_{10k}$ result, where we report the test accuracy obtained at step 1. end-to-end training. The Random column serves as a simple base-line for the effect of depth where we take the randomly initialized network and freeze the weights of all layers apart from the linear classifier. The DGL result with an asterix was training with the NTK covariance-function. We also performed DGL LEGO where the classifier was training with NLL. For full
$\text{CIFAR10}_{45k},\text{BMNIST}_{2k}$, $\text{MNIST}_{10k}$, and $\text{MNIST}_{50k}$ and got 51.85\%, 99.21\%, 97.1\%, 98.14\% test-error respectively.

\begin{table}[t]
\caption{Test accuracy of LEGO with DGL compared with other approaches}
\label{TestResults}
\vskip 0.15in
\begin{center}
\begin{small}
\begin{sc}
\begin{tabular}{lccccl}
\toprule
Dataset & L/d & End-to-end & DGL & Random \\
\midrule
$\text{CIFAR10}_{45k}$& 3/1000 & 53.1  & 53.84 & 38.12 \\
$\text{CIFAR10}_{10k}$& 3/1000 & 44.68  & 43.67* & 32.36 \\
$\text{CIFAR10}_{10k}$& 5/2000 & 45.40  & 47.45 & 34.28 \\
$\text{MNIST}_{50k}$& 2/2000 & 98.6  & 98.2 & 93.35 \\
$\text{MNIST}_{10k}$ & 2/2000 & 97.71 & 97.18  & 94.42  \\
$\text{MNIST}_{10k}$ & 3/1000 & 96.59 & 97.15  & 92.18 \\ 
$\text{BMNIST}_{2k}$ & 2/20  &  98.52 & 99.26 & 87.29 \\
$\text{BMNIST}_{2k}$ & 3/20  &  98.61 & 99.21 & 93.52 \\
\bottomrule
\end{tabular}
\end{sc}
\end{small}
\end{center}
\vskip -0.1in
\end{table}

\subsection{Discussion and outlook}
In this work we derived a correspondence between full-batch gradient descent with weight-decay and white-noise on a DNN  and Bayesian Inference on a corresponding stochastic process. This stochastic processes describes the distribution over the space of functions induced by the DNN when its weights are sampled from an iid prior dictated by the noise and weight-decay factor. Our correspondence complements the NTK correspondence [\citet{Jacot2018}] which is valid at zero noise and large width.

At large width, both correspondences along with the assumption of no co-adaptation between layers, determine a set of layer-wise loss function we dubbed the DGL loss. Layer-wise greedy training using DGLs was shown to give SOTA results on CIFAR10 and MNIST and capable of monitoring standard SGD end-to-end training. For special architectures the DGLs resemble other recent layer-wise loss function but nonetheless outperform them. In addition they generalizes trivially to CNNs by simply replacing the covariance-function with that of the CNN top-network. Notwithstanding the challenge here is to get an approximation for the CNN covariance-function [\cite{Novak2018}]. Although our DGL training is slower than SGD, approximate inference methods such as KISS-GP [\cite{Wilson2015}] may prove useful. Also we found that working with minibatches of size $1000-6000$ did not harm DGL performance on CIFAR10. 

Our goal here was to shed light on the internal representations created during training. Having a well-founded and tested analytic expression for the layer-wise loss is clearly a first step in this direction. Notwithstanding the actual role of deeper layers remains obfuscated by the large matrix inversion involved in DGL. It is highly likely that in the large dataset limit our expressions could be further simplified using equivalence kernel (EK) type results for the generalization error [\cite{Rasmussen2005}]. The latter provide simple analytic expressions in terms of the target function $g(x)$ and its projection on the eigenfunctions of the covariance-function. While a careful analysis of this is left for future work, we conjecture that the DGLs optimize $h^l(x)$ so that the function $g(h^l)$ has more support in the leading eigenvalues of the covariance-function of the top-network.


\section*{Bibliography}
\bibliography{LEGO}
\bibliographystyle{icml2019}

\end{document}


\title{The role of a layer in deep learning: a Gaussian Process perspective--- Supplemental material}

\maketitle
\section{Derivation of the DGL functions}
Here we consider a multi-label classification dataset (${\rm D}$) consisting of $N$ data points each described by a $d$ dimensional vector $x_n$ and a "one-hot“ two dimensional label (target) vector ($l = \{(1,0,...),(0,1,...),...\}$) for each class. As in \cite{Rifkin2004,Lee2018} we treat classification as a regression task where the network's outputs for a given class are optimized to be close to the one-hot label (MSE loss).  

Next we define the $n$-left-out dataset (${\rm D}_n$) consisting of all points except the point $n$. Our starting point for defining the DGL is the Bayesian prediction formula for the label vector ($l^*_n$) of an unseen datapoint ($x_n$) (unseen with respect to $\{rm D_n\}$)
\begin{align}
l^*_n &= \sum_{qp} [K({\rm D})]_{nq}[\{B({\rm D_n})\}^n]_{qp}l_p \\ \nonumber
[\{A\}^n]_{pq} &\equiv \begin{cases}
        A_{pq} \,\,\,\ \text{$p \neq n$ and $q \neq n$,}
        \\
        0 \,\,\,\,\,\,\,\,\,\ \text{otherwise}.
        \end{cases}
\end{align}
where $K({\rm D})_{pq} = K(x_p,x_q)$ is the covariance function projected on the dataset ${\rm D}$, $B({\rm D_n}) = [K({\rm D_n})+\sigma^2 I_{N-1}]^{-1}$ where $K({\rm D_n})$ is the $(n,n)$-minor of $K({\rm D})$ or equivalently the covariance-function projected onto ${\rm D_n}$, and $I_{N-1}$ is the identity matrix in an $N-1$ dimensional space. Note that we choose indices to remain faithful to data-points, so that the indices of $K({\rm D_n})$ are chosen to be the set $\{1,...,n-1,n+1,...,N\}$ rather than $\{1..N-1\}$. 

It would be convenient both analytically and numerically to relate $B({\rm D_n})$ and $B({\rm D}) = [K({\rm D})+\sigma^2 I_{N}]^{-1}$. To this end we employ a relation between inverse of a positive definite matrix ($Q$) and its $(n,n)-$minor ($Q_n$) 
\begin{align}
[Q^{-1}_n]_{pq}&= [Q^{-1}]_{pq}- \frac{[Q^{-1}]_{pn} [Q^{-1}]_{nq}}{[Q^{-1}]_{nn}} \\ \nonumber 
[\{Q^{-1}_n\}^n]_{pq}&= [Q^{-1}]_{pq}- \frac{[Q^{-1}]_{pn} [Q^{-1}]_{nq}}{[Q^{-1}]_{nn}} \\ \nonumber 
\end{align}
Notably since $Q$ is positive definite and bounded, $Q^{-1}$ is also positive definite and so the above denominator is always nonzero. Note that since $K({\rm D})$ is semi-positive-definite $K({\rm D})+\sigma^2 I$ is positive-definite. The difference on the r.h.s. of both of the above two equations lays solely in allowed values of $p,q$ ($\{1,...,n-1,n+1,...,N\}$ for the first Eq. and $\{1,...,N\}$ for second). 


Following this one can show that  
\begin{align}
\label{Eq:Relations}
K({\rm D_n})&B({\rm D_n}) = I_{N-1} - \sigma^2 B({\rm D_n}) \\ \nonumber 
& =_{\lim \sigma \rightarrow 0} I-P_{Ker[K({\rm D_n})]} = P_{Im[K({\rm D_n})]} \\ \nonumber 
[K({\rm D})&\{B({\rm D_n})\}^n]_{pq} = \delta_{pq} - \sigma^2 [B({\rm D})]_{pq} \\ \nonumber 
&- \delta_{pn} \frac{[B({\rm D})]_{nq}}{[B({\rm D})]_{nn}} + \sigma^2 \frac{[B({\rm D})]_{pn} [B({\rm D})]_{nq}}{[B({\rm D})]_{nn}}  \\ \nonumber 
&=_{\lim \sigma \rightarrow 0}  \delta_{pq}  - \delta_{pn} \frac{[B({\rm D})]_{nq}}{[B({\rm D})]_{nn}} \\ \nonumber
&- \left( P_{Ker[K({\rm D_n})]}]_{pq} - \frac{[P_{Ker[K({\rm D_n})]}]_{pn} [P_{Ker[K({\rm D_n})]}]_{nq}}{[P_{Ker[K({\rm D_n})]}]_{nn}} \right)
\end{align}
where $P_{V}$ is the projector onto the subspace $V$, $Ker[K({\rm D_n})]$ is the kernel subspace of $K({\rm D_n})$, and $Im[K({\rm D_n})]$ is the image subspace of $K({\rm D_n})$. 

Turning to the variance in the predicted target vector ($l^*_n$) the standard formula gives \cite{Rasmussen2005}
\begin{align}
K^*_{n} &= [K({\rm D})]_{nn}-\sum_{pq}  [K({\rm D})]_{n p} [\{B({\rm D_n})\}^n]_{pq} [K({\rm D})]_{q n} 
\end{align}
which using the above relations gives 
\begin{align}
K^*_{n} &= \frac{1}{[B({\rm D})]_{nn}} - \sigma^2 
\end{align}
note that since $B({\rm D})$ is positive definite with maximal eigenvalue of $1/\sigma^2$ we get that $[B({\rm D})]_{nn} < 1/\sigma^2$ and therefore the variance is non-negative as required.

We next define the DGL function as the MSE loss of the Bayesian prediction
\begin{align}
L_{DGL} &= \sum_n |l^*_n - l_n|^2 
\end{align}
Notably one can also add the variance ($\sum_n K_{nn}$) to this expression making it a more accurate measure of the expected MSE loss. For simplicity and since we found that it makes little difference in practice we did do so in the text. The Github repository we opened has this option available. In the generic case in which the covariance-matrix has no kernel and taking the limit of zero $\sigma$ we obtain 
\begin{align}
\label{Eq:DGLLoss}
L_{DGL} &= \sum_n \frac{\sum_{q,p} [B({\rm D})]_{qn}[B({\rm D})]_{np}(l_q \cdot l_p)}{[B({\rm D})]^2_{nn}}
\end{align}

\section{Contrast with Information Bottleneck approaches} 
It is interesting to compare $L_{DGL@Linear}$ with a different loss function drawn from recent works on the information bottleneck (IB) \cite{Tishby2015,Tishby2017}. In those works it was argued that the role of a layer was to compress the layers representation while maintaining the information on the labels. Formally this means minimizing the mutual information quantity 
\begin{align}
L_{IB} &= I(h;x) - \beta I(x;l)
\end{align}
for large $\beta$. A subtle yet important issue here is the fact that for deterministic networks these mutual information quantities are either constant or infinite depending how one views the entropy of a point. To overcome this the original works used binning of some linear dimension ($l_b \ll 1$) in activation space and other works added a Gaussian noise of variance $\epsilon^2$ to $h^l_n$ \cite{Saxe2018}. In case in which three data-points becoming $l_b$- or $\epsilon$-close are rare, both regularization schemes effectively lead to a pairwise interaction between data-points \cite{Kolchinsky2017,Goldfeld2018} (see also next section). Notably this is almost always the case at high dimension or when the regulator is taken to zero. For the Gaussian regulator the resulting loss is particularly simple and given by (see next section)
\begin{align}
L_{IB,\beta} &= \sum_{nm} [\beta (1-l_n l_m)-1] \Delta S_{\eta}(|h_n - h_m|) 
\end{align}
where $\Delta S_{\epsilon}(l)$ is an Gaussianly decaying interaction on the scale of $\epsilon$ given explicitly be the difference in entropy between two $d-$dimensional Gaussian distribution of variance $\epsilon^2$ and a mixture of such Gaussians as distance $l$. 

\begin{figure}[h]
	\centering
	\includegraphics*[width=8cm,trim=0 480 440 30,clip]{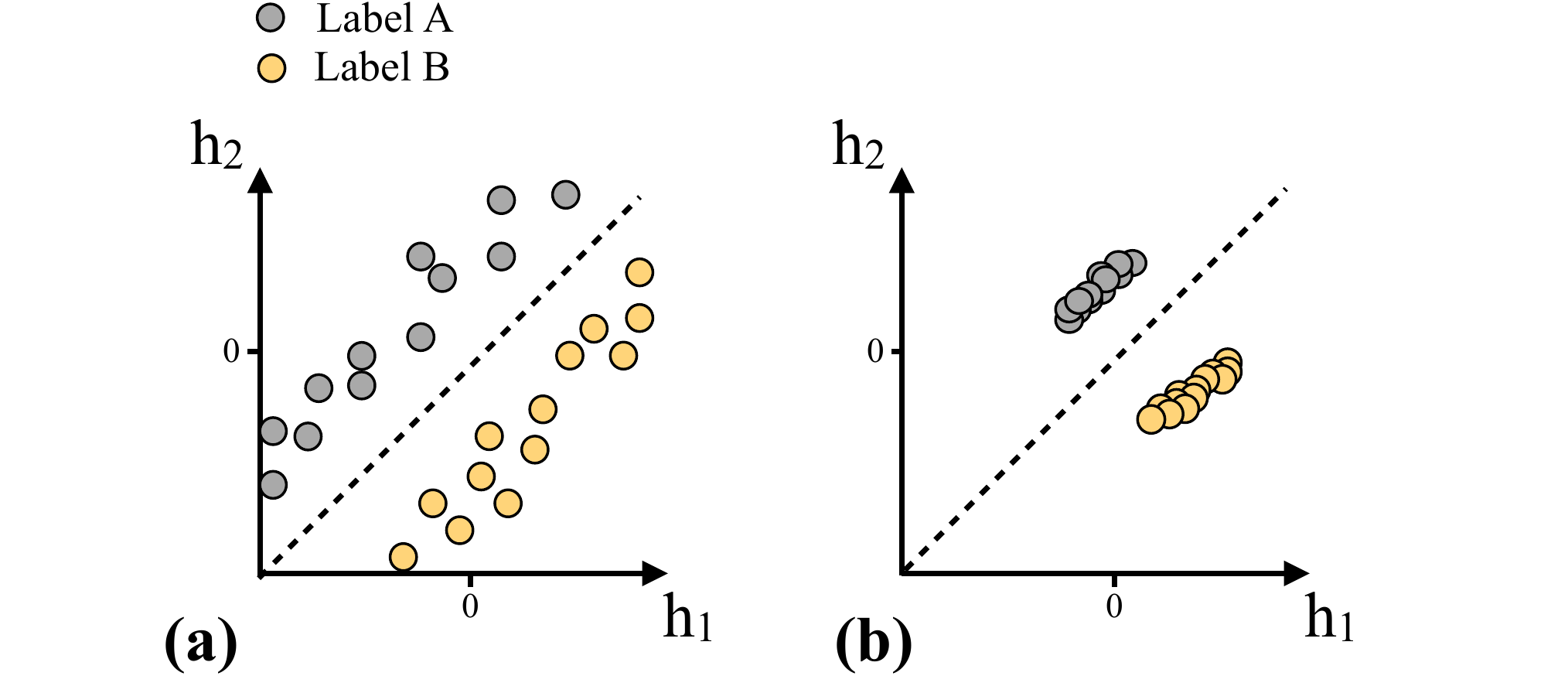}
	\caption{Two pre-classifier data representations. {\bf (a)} A typical pre-classifier dataset representation. {\bf (b)} Droplet formation encouraged by Information Bottleneck loss. The DGL before the classifier seems to capitalize on the fact that both plots are related through normalizing the axis by the co-variance matrix.   \label{Particles}} 
\end{figure}

At the input of the linear classifier, one can easily see the differences between the layer-representations favored by $L_{IB,\beta}$ and by $L_{DGL@Linear}$ (see Fig. \ref{Particles}). The former, being unaware of the classifier or the architecture, simply encourages the formation of droplets which, as argued previously, are not a faithful measure of linear separability. To achieve this unnecessary goal it is likely to compromise on the margin. The latter being aware of the classifier, encourages linear separability. We conclude that $L_{IB,\beta}$ is unlikely to be a good layer-wise loss function close to the classifier. This lack of architecture awareness of IB (regulated using binning or Gaussian noise) is generally concerning. 

\section{Information Bottleneck from the Pair Distribution Function.} The Information Bottleneck (IB) approach asserts \cite{Tishby2015,Tishby2017} that each layer, having activations $T$, minimizes the loss function $I(X:T)-\beta I(Y:T)$, where $I(X:T)$ ($I(Y:T)$) is the mutual information between the activations and the input (label) and $\beta$ is an undetermined layer specific constant which is usually order of a 100 \cite{Tishby2017}. Notably IB was proposed for deterministic network in which $T$ is a deterministic function of $X$. As commented in many works \cite{Saxe2018,Kolchinsky2017}, in such settings mutual information quantities are ill defined and require a regulator. The regulator defines how much information is in one data-point and how close two points have to be to collapse into one point. One type of regulator several authors recommend \cite{Saxe2018,AAMINE}, consists of adding a very small Gaussian random noise $\epsilon$ to $T$ and using that perturbed $T+\epsilon$ in the above loss. 

For $\epsilon$ much smaller than the typical inter-datapoint spacing and at high dimension, one can fairly assume that pairs of data-points coming $\epsilon$ close in the space of activations cause the vast majority of information loss whereas triplets of the datapoints coming $\epsilon$ close are far more rare. Clearly for low enough $\epsilon$ (i.e the deterministic limit) it would always be true unless three points happen to collapse exactly on one another. Taking this as our prescription for determining $\epsilon$, we show below that mutual information becomes a property of the pair-distribution-function (PDF) of the dataset (defined below) and as a result the IB compression can be measured only through knowledge of the pair-wise distances between all points. Such PDFs were analyzed in Ref. \cite{Goldfeld2018} and indeed compression (following auxiliary noise addition) was linked to reduction of pairwise distance in these PDFs. 

We turn to establish the mapping between mutual information with a small $\epsilon$-noise regulator and the pair-distribution function. For brevity we focus only on $I(T+\epsilon:X)$. We make the reasonable assumption that data-points ($x_n$) have no repetitions and are all equality likely. Using $I(T+\epsilon:X) = H(T+\epsilon) - H(T+\epsilon|X)$ we first find that the second contribution is just the entropy of $\epsilon$ ($H(\epsilon)$). The latter is $d-$dimensional Gaussian distribution with variance $\sigma_{\epsilon}$, which we denote by $H(\epsilon)$. The former is the entropy of $T+\epsilon$. In cases where all data-points in $T$ space ($h_n = T(x_n)$) are much further apart on the scale of $\sigma_{\epsilon}$ entropy becomes that of choosing a data-point ($\log(N)=H(X)$, where $N$ is the number of datapoints) plus that a single datapoint $H(T+\epsilon) = H(X) + H(\epsilon)$. This implies that $I(T+\epsilon:X)=H(X)$ as expected in this limit. Next consider the case where some points are far apart but some point are bounded to pairs. The entropy is now given by
\begin{align}
H(T+\epsilon:X) &= \log(N) + H(\epsilon) \\ \nonumber &+ \frac{2}{N} \sum_p [H(\Delta_p; \sigma_{\epsilon}) - H(\infty;\sigma_{\epsilon})]
\end{align}   
where $p$ runs over all pairs, $\Delta_p$ is the distance between members of the pair, and $H(\Delta_p; \sigma_{\epsilon})$ is the entropy of mixture of two d-dimensional Gaussians with variance $\sigma_{\epsilon}$ at distance $\Delta_p$. Noting that $H(\Delta; \sigma_{\epsilon}) - H(\infty;\sigma_{\epsilon})]$ decays as $e^{-\Delta^2/\sigma_{\epsilon}^2}$ one can just as well extend this sum over pairs to a sum over all points finally arriving at  
\begin{align}
H(T+\epsilon:X) &= \log(N) + H(\epsilon) \\ \nonumber &+ \frac{1}{N} \sum_{n,m} [H(t_n-t_m; \sigma_{\epsilon}) - H(\infty;\sigma_{\epsilon})]
\end{align} 

A summation of two particles/data-points terms as the one above can always be expressed using the pair-distribution-function (PDF) whose standard definition is \begin{align}
PDF_{all}(r) &= \frac{1}{N(N-1)} \sum_{n \neq m} \delta(|h_n - h_m| - r)  
\end{align} 
it is then easy to verify that 
\begin{align}
I(T+&\epsilon:X) = \log(N) \\ \nonumber &+(N-1)\int dr PDF_{all}(r)  [H(r; \sigma_{\epsilon}) - H(\infty;\sigma_{\epsilon})]
\end{align}
Similarly $I(T+\epsilon:Y)$ can be expressed using the opposite-label PDF given by 
\begin{align}
PDF_{+-}(r) &= \frac{4}{N^2} \sum_{n,m*} \delta(|h_n - h_{m*}| - r)  \\ \nonumber
I(T+\epsilon:Y) &= \log(N) \\ \nonumber &+\frac{N}{2} \int dr PDF_{+-}(r)  [H(r; \sigma_{\epsilon}) - H(\infty;\sigma_{\epsilon})]
\end{align} 
where $n$ and $m*$ scan data-points with opposite labels. We thus conclude that optimization the IB functional following $\epsilon$-noise regularization, either in the limit of $\epsilon \rightarrow 0$ or in the limit where three points reaching a distance of $\epsilon$ are rare, is simply a particular type of label dependent pairwise interaction. 

\section{DGL for the pre-classifier layer}
Here we derive in detail the DGL of pre-classifier layer. The inverse of $K_{nm} +  \sigma^2 {\rm I}$. This matrix is defined by 
\begin{align}
B_{nm} &= (\sigma_w H H^T +  \sigma^2 {\rm I})^{-1}
\end{align}
where we recall that $H$ is an $N$ by $d$ given by $[H]_{ni} = h_i(x_n) = [L_W(x_n)]_i$. Taking the limit of $\sigma \rightarrow 0$ one immediately has that 
\begin{align}
B_{nm} &= \frac{1}{\sigma^2} P_{Ker[H H^T]} + O(\sigma^0)
\end{align}
Without fine tuning $\Sigma = H^T H$ is positive-definite. Notably this statement is equivalent to saying that the $N \times d$ matrix $H$ has $d$ linearly independent columns. Notably when $N \geq d$ having two linearly dependent coloumns requires fine-tunning of $N-d+1$ parameters, hence when $N \gg d$ this becomes extremely unlikely under any reasonable ensemble for $H$. 

In this case one can show that $P_{Ker[H H^T]}=I-H \Sigma^{-1} H^T$. Indeed 
\begin{align}
&P_{Ker[H H^T]}^2 = I - 2 H \Sigma^{-1} H^T + H \Sigma^{-1} H^T H \Sigma^{-1} H^T \\ \nonumber 
&= I - 2 H \Sigma^{-1} H^T + H \Sigma^{-1} H^T \\ \nonumber 
&= P_{Ker[HH^T]} \\ 
&H H^T P_{Ker[H H^T]} = 0  \\ 
&[I-P_{Ker[H H^T]}] H H^T = H H^T
\end{align}
This equation implies that $P_{Ker[H H^T]}$ is a projector (in fact an Hermitian projector as is easy to verify). The second that its image is in the kernel of $H H^T$. The third that its kernel is in the image of $H H^T$. All in all it implies that it is a projector whose image coincides with the kernel of $H H^T$ as required. 

Next we consider Eqs. (\ref{Eq:Relations}). The fact that the kernel is non-trivial adds several complicated terms to our loss. These all term depend on $[P_{Ker[HH^T]}]_{pq}$ which we next expand as 
\begin{align}
[P_{Ker[HH^T]}]_{pq} &= \delta_{pq} - [P_{Im[HH^T]}]_{pq} = \delta_{pq} + O(d/N). 
\end{align}
we in the right hand side we noted that $Im[HH^T]$, the image of $HH^T$, is of dimension $d \ll N$, consequently the norm of the operator $[P_{Im[HH^T]}]_{pq}$ is $d$, while the norm of the $\delta_{pq}$ $N$. Notably this statement is only accurate element-wise when we assume that $P_{Im[HH^T]}$ has no particular relation with the basis on which the matrix is written on. For this not to hold it would require that at least one $d-$dimensional row of $H$ is orthogonal to all the remaining $N-1$ rows. This is again exponentially unlikely in the limit of $N \gg d$ under any reasonable ensemble for $H$. 

Accordingly we treat the expansion in the $d/N$ as an expansion in $P_{Im[HH^T]} = H \Sigma^{-1} H^T$. For instance we can then expand 
\begin{align}
&\sum_n \frac{[B({\rm D})]_{pn}[B({\rm D})]_{nq}}{[B({\rm D})]^2_{nn} } = \sum_n \frac{[B({\rm D})]_{pn}[B({\rm D})]_{nq}}{1+O(d/N)} \\ \nonumber 
&= P_{Ker[HH^T]} + O\left((d/N)^2\right)
\end{align}
Plugging this into Eq. (\ref{Eq:DGLLoss}) we obtain 
\begin{align}
L_{DGL@Linear} &= \sum_{n}|l_n|^2 -\sum_{nm} [H \Sigma^{-1}H^T]_{nm} (l_n \cdot l_m) 
\end{align}
as in the main text.

\section*{Bibliography}
\bibliography{LEGO}
\bibliographystyle{icml2019}